\documentclass[10pt,twocolumn,letterpaper]{article}

\usepackage{cvpr}
\usepackage{times}
\usepackage{epsfig}
\usepackage{graphicx}
\usepackage{amsmath}
\usepackage{amssymb}
\usepackage{caption}
\usepackage{tikz}
\usepackage{booktabs}
\usepackage{graphicx}
\usepackage{caption}
\usepackage{subcaption}
\usepackage{float}
\usepackage{footnote}

\newcommand\blfootnote[1]{%
  \begingroup
  \renewcommand\thefootnote{}\footnote{#1}%
  \addtocounter{footnote}{-1}%
  \endgroup
}

\usepackage[pagebackref=true,breaklinks=true,letterpaper=true,colorlinks,bookmarks=false]{hyperref}

\cvprfinalcopy 

\def\cvprPaperID{****} 

\ifcvprfinal\fi
\begin{document}

\title{DinoTwins: Combining DINO and Barlow Twins for Robust, Label-Efficient Vision Transformers}

\author{Podsiadly, Michael \quad Lay, Brendon K\\\\
Georgia Institute of Technology\\
225 North Avenue NW, Atlanta, GA 30332\\
{\tt\small \{mpodsiadly3, blay7\}@gatech.edu}
}

\newcommand\imgHeight{1in}
\newcommand\imgSpacing{0.15\textwidth}

\twocolumn[{%
\renewcommand\twocolumn[1][]{#1}%
\maketitle
\begin{center}
  \centering
    \captionsetup{type=figure}
    \begin{subfigure}[t]{\imgSpacing}
        \centering
        \includegraphics[height=\imgHeight]{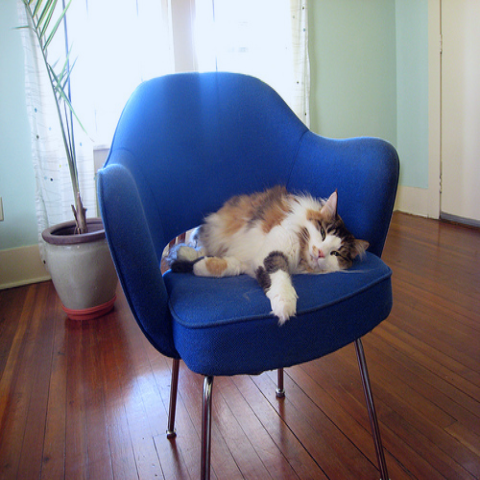}
    \end{subfigure}%
    ~
    \begin{subfigure}[t]{\imgSpacing}
        \centering
        \includegraphics[height=\imgHeight]{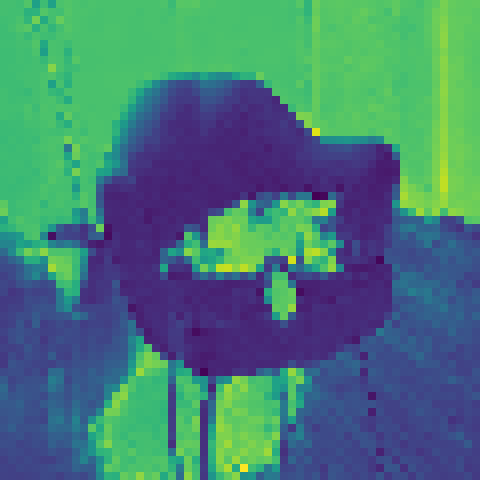}
    \end{subfigure}%
    ~
    \begin{subfigure}[t]{\imgSpacing}
        \centering
        \includegraphics[height=\imgHeight]{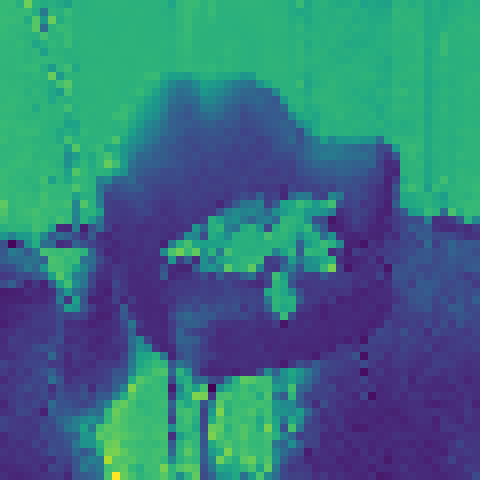}
    \end{subfigure}%
    ~
    \begin{subfigure}[t]{\imgSpacing}
        \centering
        \includegraphics[height=\imgHeight]{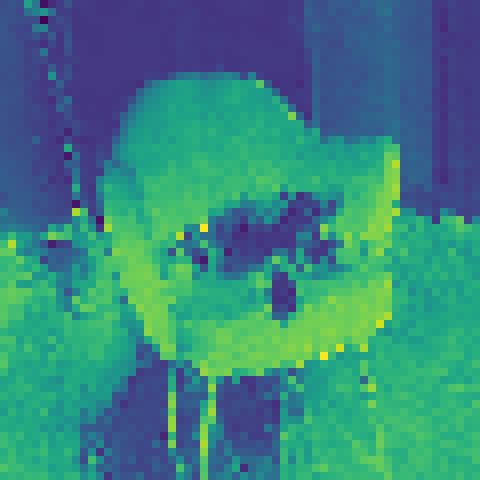}
    \end{subfigure}%
    \vskip\baselineskip 
    \begin{subfigure}[t]{\imgSpacing}
        \centering
        \includegraphics[height=\imgHeight]{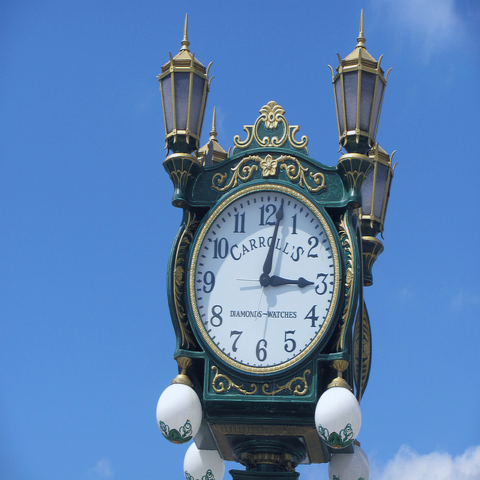}
    \end{subfigure}%
    ~
    \begin{subfigure}[t]{\imgSpacing}
        \centering
        \includegraphics[height=\imgHeight]{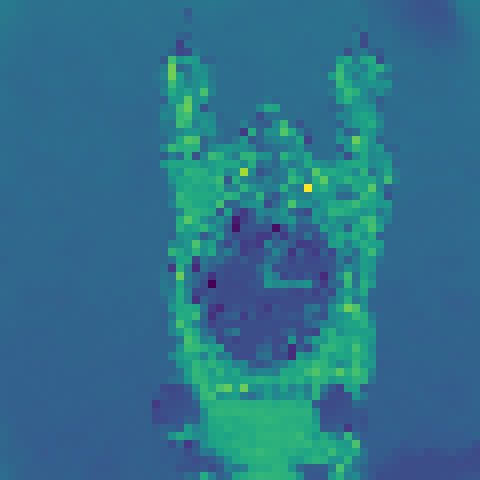}
    \end{subfigure}%
    ~
    \begin{subfigure}[t]{\imgSpacing}
        \centering
        \includegraphics[height=\imgHeight]{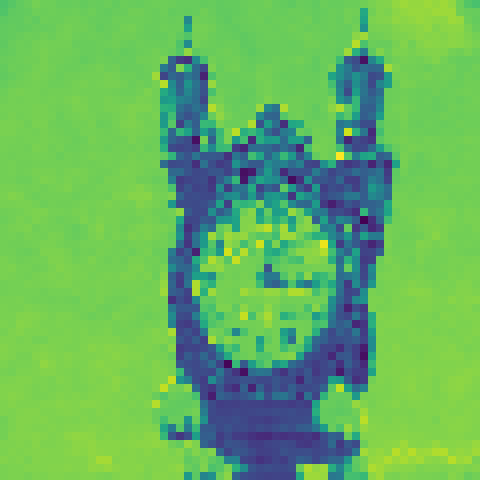}
    \end{subfigure}%
    ~
    \begin{subfigure}[t]{\imgSpacing}
        \centering
        \includegraphics[height=\imgHeight]{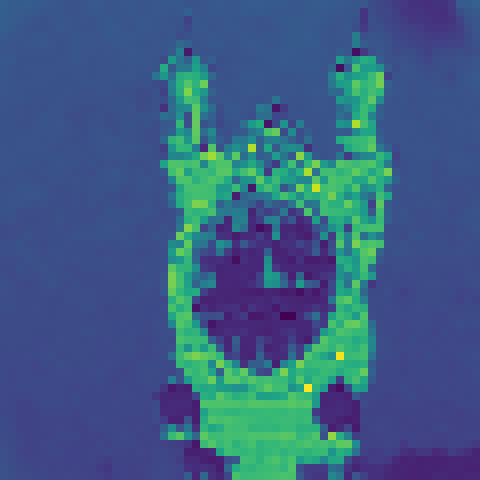}
    \end{subfigure}%
    \captionof{figure}{\textbf{Self-attention maps from the final layer of a Vision Transformer trained with three different methods: Barlow Twins, DINO, and DINO + Barlow Twins.} Each map shows the attention of the [CLS] token using 8×8 patches. Despite the absence of labels or supervision, the models learn class-specific features, demonstrating emergent object segmentation capabilities under self-supervised training.}
    \label{fig:vis_atten}
\end{center}%
}]

\blfootnote{
Code: \url{https://github.gatech.edu/mpodsiadly3/DinoTwins.git}
}

\begin{abstract}
Training AI models to understand images without costly labeled data remains a challenge. We combine two techniques—DINO (teacher-student learning) and Barlow Twins (redundancy reduction)—to create a model that learns better with fewer labels and less compute. While both DINO and Barlow Twins have independently demonstrated strong performance in self-supervised learning, each comes with limitations—DINO may be sensitive to certain augmentations, and Barlow Twins often requires batch sizes too large to fit on consumer hardware. By combining the redundancy-reduction objective of Barlow Twins with the self-distillation strategy of DINO, we aim to leverage their complementary strengths. We train a hybrid model on the MS COCO dataset using only 10\% of labeled data for linear probing, and evaluate its performance against standalone DINO and Barlow Twins implementations. Preliminary results show that the combined approach achieves comparable loss and classification accuracy to DINO while maintaining strong feature representations. Attention visualizations further suggest improved semantic segmentation capability in the hybrid model. This combined method offers a scalable, label-efficient alternative for training ViTs in resource-constrained environments.
\end{abstract}

\section{Introduction}\label{introduction}
Labeling data often demands specialized expertise—for example, annotating medical imaging data requires trained professionals. Our method could reduce labeling costs while maintaining accuracy. Vision Transformers (ViTs) have emerged as powerful architectures for visual representation learning, but they are notoriously data-hungry and computationally expensive when trained in a fully supervised manner. Self-supervised learning (SSL) frameworks like \textbf{DINO}\cite{DINO} and \textbf{Barlow Twins}\cite{BarlowTwins} offer promising alternatives by eliminating the need for labeled data while still enabling ViTs to learn semantically meaningful features. This paper aims to explore the combination of these two innovations, as no prior work has integrated DINO’s semantic clustering with Barlow Twins’ feature decorrelation in the context of Vision Transformers. However, each of these methods comes with its own strengths and weaknesses.

\section{Background}
\subsection*{Barlow Twins: Redundancy Reduction}
Barlow Twins is a self-supervised learning framework designed to extract robust and diverse feature representations from images by reducing redundancy between learned features. It operates on the principle that good representations should be invariant to augmentations while avoiding feature collapse. To achieve this, it processes two differently augmented views of the same image through identical encoder networks, resulting in embedding vectors $\mathbf{Z}_A$ and $\mathbf{Z}_B$. These embeddings are batch-normalized and used to compute a cross-correlation matrix $\mathbf{C}$.

The Barlow Twins loss minimizes the deviation of this matrix from the identity, encouraging diagonal elements to be close to 1 (i.e., the same features across views) and off-diagonal elements to be close to 0 (i.e., decorrelated features). This dual objective ensures that features are not only invariant to image augmentations but also non-redundant. 

An important aspect of this formulation is that it avoids representation collapse—a common failure mode in self-supervised learning—without requiring negative pairs, unlike contrastive methods. Furthermore, decorrelated embeddings can improve the downstream usefulness of features by ensuring that each latent dimension captures a unique aspect of the input.

Let $\mathbf{Z}_A, \mathbf{Z}_B \in \mathbb{R}^{N \times D}$\cite{BarlowTwins} be the embeddings from two augmented views of the same batch of images, produced by two identical networks. The embeddings are batch-normalized and used to compute the cross-correlation matrix:

\begin{equation}
\mathbf{C}_{ij} = \frac{1}{N} \sum_{n=1}^N z_{A,ni} \cdot z_{B,nj}
\end{equation}

The Barlow Twins loss is then defined as:

\begin{equation}
\mathcal{L}_{\text{BT}} = \sum_{i} (1 - \mathbf{C}_{ii})^2 + \lambda \sum_{i} \sum_{j \neq i} \mathbf{C}_{ij}^2
\end{equation}

The first term encourages invariance (diagonal elements close to 1), while the second term enforces redundancy reduction by minimizing the off-diagonal elements of the cross-correlation matrix.

\subsection*{DINO: Self-Distillation without Labels}

\textit{DINO} (Distillation with No Labels) takes a different approach by using a teacher--student setup.\cite{DINO} Both networks process differently augmented views of the same image. The student learns to match the teacher’s output distribution over image tokens, including the special [CLS] token.

Given a batch of images, the student network $f_s$ and teacher network $f_t$ produce softmax-normalized outputs $\mathbf{p}_s$ and $\mathbf{p}_t$\cite{DINO}, respectively:

\begin{equation}
\mathbf{p}_s = \text{softmax}\left( \frac{f_s(x)}{\tau_s} \right), \quad
\mathbf{p}_t = \text{softmax}\left( \frac{f_t(x')}{\tau_t} \right)
\end{equation}

The DINO loss is a cross-entropy between the two distributions:

\begin{equation}
\mathcal{L}_{\text{DINO}} = - \sum_{k=1}^{K} p_{t}^{(k)} \log p_{s}^{(k)}
\end{equation}

Here, $x$ and $x'$ are two augmentations of the same image, and $\tau_s, \tau_t$ are temperature parameters that control distribution sharpness. The teacher network is updated using an exponential moving average (EMA) of the student weights.\cite{DINO}

\subsection*{Motivation for Integration}

We hypothesize that combining the invariance and decorrelation objective of Barlow Twins with the semantic clustering and global attention behavior of DINO can lead to representations that are more robust to image transformations and generalize better. To this end, we propose a hybrid framework that replaces DINO’s cross-entropy objective with the Barlow Twins loss function while maintaining the overall DINO architecture. The combined objective is expressed as:

\begin{equation}
\mathcal{L}_{\text{Hybrid}} = \mathcal{L}_{\text{BT}} + \alpha \cdot \mathcal{L}_{\text{DINO}}
\end{equation}

where $\alpha$ is a hyperparameter balancing the influence of semantic alignment and redundancy reduction. This hybrid loss seeks to combine local decorrelation with global semantic consistency, potentially leading to better representations for downstream classification and interpretability.

\section{Method}
Our experiment consisted of training a Vision Transformer simultaneously using Barlow Twins and DINO. We hypothesized that these two loss functions are complimentary and that by combining them the best aspects of both could be brought into training protocol. Specifically, we expected to have the feature decorrelation and efficient training of Barlow Twins along with the semantic consistency and spatial awareness of DINO, yielding richer and more transferable representations. 

Following the data augmentation protocols used in original papers, the training pipeline begins by applying heavy data augmentation--random crops, color jitter, Guassian blur, and solarization--to produce two sets of training data. The first set consists of two views of the same image and is used for computing the Barlow Twins loss. The second consists of two global and six local views of each image to be used for DINO's self-distillation loss. For Barlow Twins, the distance between the cross-correlation matrix between paired embeddings and the identity matrix was computed. For DINO, the global and local crops were fed through the teacher and student models and the distance between the weights of the student model and those of the teacher model was computed. As in the DINO paper, the teacher's weights were updated as an exponential moving average of the student's weights and centering and sharpening steps were used to avoid collapse. These two losses were added (scaling Barlow Twins appropriately) before being backpropagated. 

To ensure a fair comparison, we trained two additional Vision Transformer models, one using only Barlow Twins and the other only DINO. After 100 epochs, each pretrained model was frozen, and a linear classification head was appended and trained using only 10\% of the labeled CIFAR-10 dataset. Additionally, class activation maps of the the models' attention layers were examined to see if the models had learned semantically meaningful features.



\subsection{Datasets}

We used a subset of the MS COCO 2017 dataset\cite{Coco}, which contains approximately 118,000 labeled training images and 41,000 test images across 80 object categories. For our experiments, we randomly sampled 256 images from the training set and discarded all labels, using them solely for self-supervised feature learning. 

Our ViT models were trained using a hybrid DINO–Barlow Twins objective, with a batch size of 8, on a single NVIDIA GeForce RTX 3080 GPU with 12GB of VRAM. During training, no class information was provided—the models learned purely from augmentations of the input images.

To evaluate the quality of the learned representations, we froze the encoder and extracted features from the final layer (i.e., the [CLS] token) without training any downstream classifier. These extracted features were then used in a separate linear evaluation protocol to assess performance, without fine-tuning or using 10\% of labeled data.

\subsection{Supervised Evaluation}

To assess the quality of the learned representations, we performed a supervised evaluation using a linear classifier. Although our primary training objective was self-supervised, we validated the representations by extracting features from each architecture and training a lightweight linear classifier on the CIFAR-10 dataset. This evaluation allowed us to benchmark the effectiveness of our DINO, Barlow Twins, and DINO+Barlow models in a downstream classification task. We referenced publicly available code from the official DINO repository by Facebook\cite{DINO_repo} to guide our linear evaluation setup.

Additionally, we visualized self-attention maps from the last layer of our Vision Transformers using both 8$\times$8 and 16$\times$16 patch sizes. These visualizations highlight the spatial attention learned by the [CLS] token in the absence of supervision. Two examples of these maps are shown in Figure~\ref{fig:vis_atten}. Visualization code was adapted from the official DINO implementation\cite{DINO_repo}.

We encountered several hardware constraints during training. Specifically, increasing the number of training samples or the batch size led to out-of-memory (OOM) errors on our NVIDIA GeForce RTX 3080 GPU with 12GB VRAM. To mitigate this, we reduced the batch size to 8 while maintaining a resolution of 224$\times$224 pixels for each image. Although lowering the input resolution to 128$\times$128 could have saved memory, it would likely reduce performance by discarding high-frequency spatial details. Alternatively, hosted GPU services such as AWS or Google Cloud were considered, but were excluded due to budget constraints.




\begin{figure}
    \centering
    \includegraphics[width=0.45\textwidth]{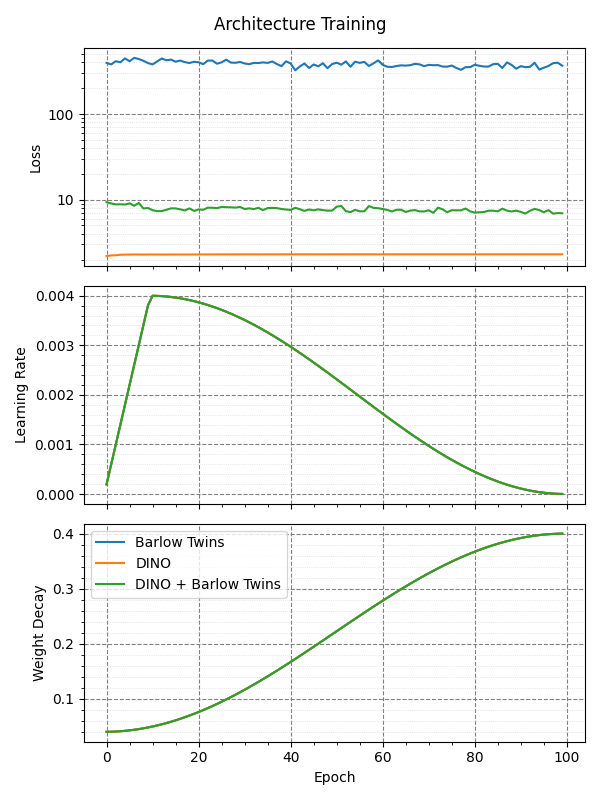}
    \caption{Training loss over 100 epochs for each self-supervised method: Barlow Twins (blue), DINO (orange), and DINO + Barlow Twins (green). While Barlow Twins shows significantly higher absolute loss values due to its correlation-based objective, all models exhibit stable convergence. The hybrid model retains stability while jointly optimizing both objectives.}
    \label{fig:training_loss}
\end{figure}






\section{Results}

We evaluated the effectiveness of Barlow Twins, DINO, and the hybrid DINO + Barlow Twins model across both quantitative metrics and qualitative visualizations.

\subsection{Training Loss and Efficiency}

Figure~\ref{fig:training_loss} shows the training loss over 100 epochs for all three models. While the Barlow Twins model displayed the highest absolute loss throughout training (averaging around 384.4), both DINO and the combined model achieved much lower losses—2.3 and 7.7 respectively (Table~\ref{tab:training_loss}). This discrepancy in scale is expected due to differences in loss formulations. Importantly, all models showed stable convergence without signs of collapse, indicating successful representation learning despite their unsupervised nature.

In terms of training time, the hybrid model incurred the highest computational cost, requiring 1,304 seconds, followed by DINO (1,127s) and Barlow Twins (778s). This reflects the added complexity of jointly optimizing both loss functions.

\begin{table}[h]
\centering
\caption{Final training loss and total training time for each model.}
\label{tab:training_loss}
\begin{tabular}{lccc}
\toprule
\textbf{Implementation} & \textbf{Mean Loss} & \textbf{Time (s)} \\
\midrule
Barlow Twins & 384.425 & 778.05 \\
DINO         & 2.297   & 1,127.01 \\
DINO + Barlow Twins & 7.720 & 1,304.07 \\
\bottomrule
\end{tabular}
\end{table}

\subsection{Linear Evaluation}

To evaluate representation quality, we trained a linear classifier on top of the frozen features extracted from each model using the CIFAR-10 dataset. As shown in Table~\ref{tab:linear_eval}, DINO and DINO + Barlow Twins performed nearly identically, achieving Top-1 accuracy of 28.88\% and Top-5 accuracy of 80.86\%. Barlow Twins trailed slightly in Top-1 accuracy (28.51\%), but matched or slightly outperformed DINO in Top-5 (80.89\%).

These results suggest that while Barlow Twins contributes to robust feature learning, the inclusion of DINO’s self-distillation mechanism offers a more consistent improvement in classification performance. The hybrid model preserved DINO’s performance, implying that the redundancy-reduction objective did not degrade semantic representation quality.

\begin{table}[h]
\centering
\caption{Linear evaluation results on CIFAR-10 using extracted [CLS] token features.}
\label{tab:linear_eval}
\begin{tabular}{lccc}
\toprule
\textbf{Implementation} & \textbf{Loss} & \textbf{Acc@1} & \textbf{Acc@5} \\
\midrule
Barlow Twins & 1.973 & 28.51\% & 80.89\% \\
DINO         & 1.974 & 28.88\% & 80.86\% \\
DINO + Barlow Twins & 1.974 & 28.88\% & 80.86\% \\
\bottomrule
\end{tabular}
\end{table}

\subsection{Attention Map Visualization}

Figure~\ref{fig:vis_atten} shows self-attention maps from the final transformer layer for each model, highlighting the [CLS] token’s attention over the input space. Even without supervision, all models developed the ability to localize salient regions of the image.

Qualitatively, DINO and DINO + Barlow Twins produced sharper, more focused attention patterns around object boundaries. In contrast, the Barlow Twins model exhibited more diffuse, less localized attention. This aligns with the expectation that DINO’s distillation framework promotes spatial consistency and high-confidence clustering. The attention behavior in the hybrid model closely matched DINO, suggesting that the addition of the Barlow Twins loss did not compromise global attention structure.

\subsection{Summary}

Overall, the DINO + Barlow Twins hybrid model maintained the strong performance and attention quality of DINO while incorporating redundancy-reduction through the Barlow Twins objective. While it did not outperform DINO in classification accuracy, it demonstrated no regression in quality, validating our hypothesis that combining these methods can produce semantically consistent and robust representations.

\begin{figure}
    \centering
    \includegraphics[width=0.45\textwidth]{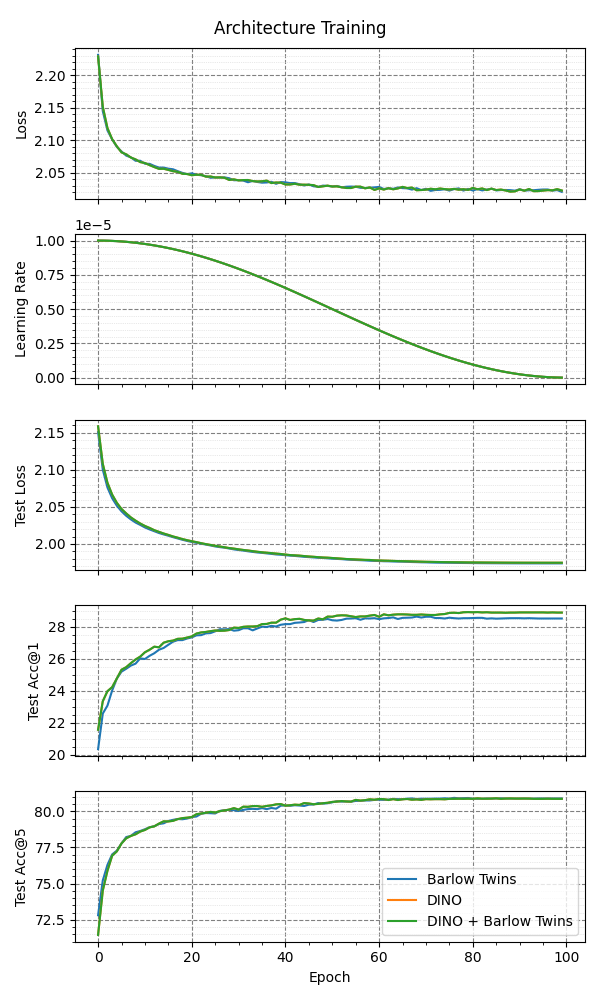}
    \caption{Random projection of extracted [CLS] token features for each model. Despite being trained without labels, all methods exhibit meaningful separation in feature space. The DINO and DINO + Barlow Twins models show tighter clustering, suggesting stronger semantic grouping than the Barlow Twins baseline.}
    \label{fig:eval_linear}
\end{figure}



\section{Discussion}

Our experimental results highlight the strengths and limitations of combining the Barlow Twins and DINO self-supervised learning frameworks. The hybrid model successfully retained the semantic consistency and spatial attention properties of DINO, while incorporating Barlow Twins’ redundancy reduction objective without degrading performance. This suggests that the objectives are at least partially compatible and can be trained jointly without adverse effects.

However, the hybrid approach did not yield significant improvements in downstream classification accuracy compared to DINO alone. This may be due to the relatively small scale of our dataset (256 samples) or the lack of careful tuning of the loss weighting hyperparameter $\alpha$. Additionally, the Barlow Twins loss exhibited a much higher absolute scale, which could have dominated the optimization unless appropriately scaled. Although loss convergence was stable, this mismatch in scale and behavior might have diluted the individual benefits of each objective.

Qualitative attention visualizations demonstrated that the hybrid and DINO models produced sharper, more localized attention patterns than the Barlow-only model. This reinforces the idea that DINO’s distillation-based training yields representations with stronger object-level semantics. The Barlow-only model, while capable of learning invariant and decorrelated features, lacked the fine-grained focus necessary for semantic object localization.

These findings support our initial hypothesis: integrating DINO with Barlow Twins does not harm the ViT’s capacity to learn semantically rich features and may enhance robustness to transformations, though further testing is needed to isolate those effects more clearly.

\section{Future Work}

This project opens several promising directions for continued exploration:

\begin{itemize}
    \item \textbf{Scale up training:} Using a larger subset of MS COCO (or full ImageNet) would better test the limits of the hybrid model’s performance, especially in low-label or zero-label regimes.
    
    \item \textbf{Loss weighting analysis:} Carefully tuning the relative weight $\alpha$ between the Barlow Twins and DINO losses could better balance decorrelation and semantic clustering, potentially improving feature quality.
    
    \item \textbf{k-NN and clustering evaluation:} In addition to linear probing, evaluating learned representations using non-parametric methods like k-NN classification or clustering metrics could reveal deeper insights into feature geometry.
    
    \item \textbf{Augmentation sensitivity:} One motivation behind this work was to improve robustness to image transformations. Systematically testing performance under various augmentation types (e.g., occlusion, viewpoint change, lighting shift) would validate this aspect of the hypothesis.
    
    \item \textbf{Compute optimization:} Training on consumer-grade GPUs imposed strict limits on batch size and resolution. Future work could leverage mixed precision training, model distillation, or cloud computing resources to explore more ambitious scales.
    
    \item \textbf{Alternative architectures:} Testing this hybrid loss on different backbone architectures (e.g., ResNets, Swin Transformers) would help assess the generalizability of the approach across vision models.
\end{itemize}

Ultimately, combining orthogonal self-supervised objectives remains a promising direction for advancing data-efficient learning in vision transformers. With further refinement, the hybrid DINO + Barlow Twins approach could provide a robust foundation for representation learning in real-world, low-label settings.

{\small
\bibliographystyle{ieee_fullname}
\bibliography{egbib}
}

\end{document}